\documentclass[letterpaper, 10 pt, conference]{ieeeconf}  

\IEEEoverridecommandlockouts                              

\usepackage{bm}
\usepackage{amsmath} 
\usepackage{graphicx} 
\let\labelindent\relax
\usepackage{enumitem}

\newlist{contrib}{enumerate}{1}
\setlist[contrib]{%
  label=\arabic*) ,      
  leftmargin=*,          
  labelsep=0.5em,        
  align=left,            
  itemsep=0.25em, topsep=0.3em
}

\title{\LARGE \bf
Dual-Mode Magnetic Continuum Robot for Targeted Drug Delivery
}

\author{Wendu Zhang, Heng Wang, Shuangyi Wang, Yuanrui Huang*
\thanks{*This work was supported by the XJTLU Research Development Fund, under RDF-24-01-005. (Corresponding Author: Yuanrui Huang)}
\thanks{Wendu Zhang is now with the Department of Mechanical and Automation Engineering, The Chinese University of Hong Kong, Hong Kong SAR, China
        {\tt\small 1155247302@link.cuhk.edu.hk}}
\thanks{Yuanrui Huang is with the School of Robotics, Xi'an Jiaotong-Liverpool University, Suzhou, China
        {\tt\small yuanrui.huang@xjtlu.edu.cn}}
\thanks{Heng Wang is with the Shien-Ming Wu School of Intelligent Engineering, South China University of Technology, Guangzhou 511442, China
        {\tt\small wanghengscut@scut.edu.cn}}
\thanks{Shuangyi Wang is with the State Key Laboratory of Multimodal Artiffcial Intelligence Systems, Institute of Automation, Chinese Academy of Sciences, Beijing 100190, China
        {\tt\small shuangyi.wang@ia.ac.cn}}
}

\begin{document}

\maketitle
\thispagestyle{empty}
\pagestyle{empty}

\begin{abstract}

Magnetic continuum robots (MCRs) enable minimally invasive navigation through tortuous anatomical channels, yet axially magnetized designs have largely been limited to bending-only motion. To expand deformation capabilities, this paper presents a simple assembly that embeds permanent magnets radially within the catheter wall, allowing a single externally steered permanent magnet to independently induce either bending or torsion. A physics-based formulation together with finite-element analysis establishes the actuation principles, and benchtop experiments validate decoupled mode control under practical fields. Building on this, a dual-layer blockage mechanism consisting of outer grooves and inner plates leverages torsional shear to achieve on-demand drug release. Finally, an in-phantom intervention experiment demonstrates end-to-end operation: lumen following by bending for target approach, followed by twist-activated release at the site. The resulting compact, cable-free platform combines versatile deformation with precise payload delivery, indicating strong potential for next-generation, site-specific therapies.

\end{abstract}

\section{INTRODUCTION}
Minimally invasive interventional techniques enable surgical instruments to reach deep-seated lesions and deliver therapeutic agents locally at the target site, thereby enhancing treatment efficacy and minimizing systemic side effects. Within this context, precise and controllable drug delivery has become a central goal in the advancement of medical robotic systems.
\par
To date, research on drug delivery in medical robotics has predominantly focused on micro/nanorobotic platforms. However, conventional nanocarriers are typically passive, lacking on-board actuation or navigation capabilities. As a result, they rely heavily on systemic circulation and increased dosing to reach therapeutic levels at the target site—raising the risk of off-target exposure and associated toxicity [1,2]. Recent progress in nanotechnology and microfabrication has led to the development of medical micro/nanorobots capable of active locomotion (often under external magnetic, acoustic, optical, or electric fields), targeted navigation, and programmable, on-demand cargo release. These features allow for spatiotemporally controlled delivery tailored to specific clinical requirements. Nonetheless, translating such systems to in vivo applications remains challenging. Their miniature and independent nature necessitates sufficient thrust generation in complex physiological environments, reliable tracking and control, and strict compliance with biocompatibility, biodegradability, and system-level coordination—including collective swarm behaviors [3–7].
\par
In contrast, magnetic continuum robots (MCRs) offer a more robust and scalable approach to locomotion and navigation by leveraging externally applied magnetic fields, thereby eliminating the need for on-board propulsion systems. Due to their inherent flexibility and ability to navigate through narrow and tortuous anatomical pathways, MCRs have emerged as a promising platform for minimally invasive interventions. By embedding magnetic materials along the robot body and steering them using controlled external magnetic fields, MCRs achieve wireless actuation within the human body [8–11]. This approach circumvents the complexity of in vivo actuators found in tendon-driven [12,13], hydraulic [14,15], or pneumatic systems [16,17], and has garnered increasing attention in recent years.
\par
Despite their potential, studies on drug delivery mechanisms using MCRs remain limited. Recent research has primarily focused on enhancing bending capabilities—typically by optimizing the spatial distribution of magnetic moments to improve bending angles and reachability for deep-seated sites [18–26]. However, this emphasis on bending has left other deformation modes underexplored. In particular, the lack of systematic investigation into torsional actuation has hindered the multifunctionality of MCRs and constrained their broader applicability in complex clinical scenarios.
\par
To address this gap, this study proposes a dual-deformation-mode MCR that simultaneously supports both locomotion and drug delivery functions. Mode switching is achieved solely by reorienting an external magnetic field using a six-degree-of-freedom (6-DoF) robotic mechanism to manipulate an external permanent magnet (EPM) [27–31].
\par
\noindent The main contributions of this study are as follows:
\begingroup\setlength{\emergencystretch}{1em}
\begin{contrib}
\item \text{Dual-Mode Magnetic Actuation for Continuum Robot:}
We present a simple yet effective design and actuation strategy that enables a single continuum robot to achieve both bending and torsion through magnetic control. 

\item \text{Twist-Triggered Drug Release Mechanism:}
We introduce a compact, dual-layer blockage structure comprising outer grooves and inner plates. This mechanism utilizes torsional shear strain to open or close the drug outlet, enabling on-demand, site-specific drug release without requiring additional delivery lines or valve components.
\end{contrib}
\endgroup
\begin{figure*}[!t]
    \centering
    \includegraphics[width=1.0\linewidth]{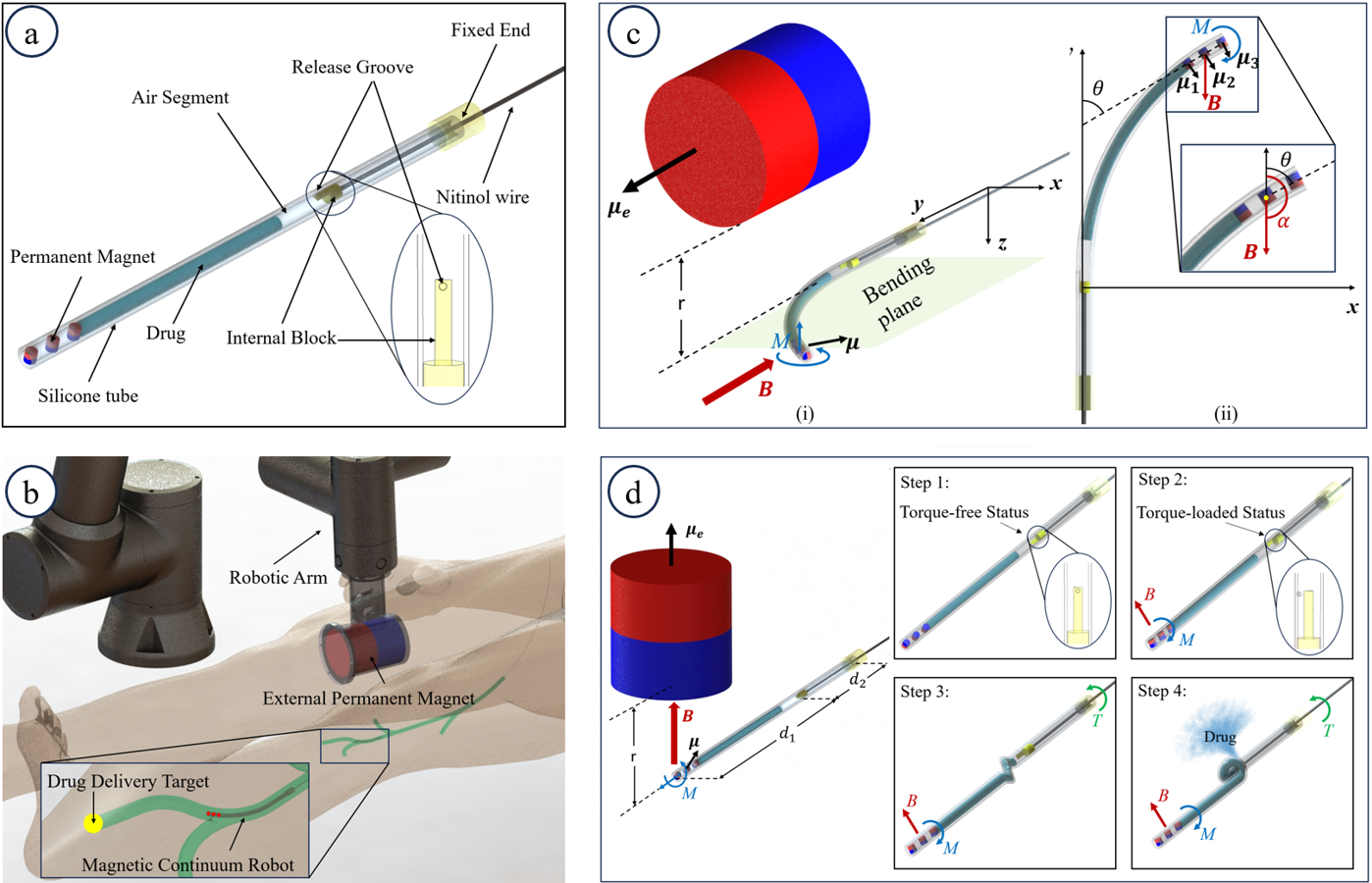}
    \caption{(a) Prototype of the dual-mode MCR (b) Overview of the proposed magnetic actuation system (c) Actuation schematic for bending mode (d) Actuation schematic for torsional mode.}
    \label{fig:1}
\end{figure*}
\par
This paper is organized as follows: Section II presents the actuation principles and control strategies for both bending and torsion, as well as the proposed drug-release mechanism. Section III details the methods used to evaluate deformation performance and presents corresponding results, including navigation and targeted drug delivery experiments conducted in a PDMS-based phantom. Section IV provides further analysis and discussion of the experimental outcomes. Finally, Section V concludes the paper and outlines future research directions.

\section{System Design and Magnetic Actuation}

MCRs achieve deformation primarily via magnetic torque, exerted on the internal permanent magnets (IPMs) by the magnetic field produced by a much larger EPM. This relationship is given by:
\begin{equation} 
\bm{m} = \boldsymbol{\mu} \times \mathbf{B}
\end{equation}
where $\boldsymbol{\mu}$ denotes the magnetic moment of the IPM, and $\bm{m}$ represents the torque exerted on the IPM by the magnetic field $\mathbf{B}$.
\par
Previous studies have embedded permanent magnets axially, aligning their magnetic moments with the central axis of the robot. This configuration ensures that when an external magnetic field is applied, it will always generate a bending moment perpendicular to the MCR’s central axis, causing it to bend while preventing other deformation modes.
\par
To enhance the deformation diversity of the MCR, this study introduces a novel design that incorporates multiple permanent magnets radially embedded at the distal tip of the catheter. A conceptual illustration is shown in Fig. 1(a). These IPMs are fixed in place by filling the spaces between them with solid material, ensuring their orientations and relative spacing remain stable during operation. This arrangement enables the magnets to generate magnetic torque in various directions under different external magnetic field configurations. Specifically, when the generated magnetic torque is misaligned with the catheter’s central axis, it induces bending; when the torque is aligned with the axis, it produces torsion. By manipulating the external magnetic field, the MCR can switch between these two deformation modes.
\par
This dual-mode deformation capability further equips the MCR to function as a drug-delivery device. It can transport drugs to specific targets and release them through a customized mechanism activated by the torsion of the catheter. Immediately proximal to the tip, the catheter is partially loaded with the drug, while an air segment is intentionally preserved between the drug and a dual-layer blockage structure. This air segment acts as a buffer, preventing unintended release until the appropriate torsional actuation is applied. A close-up of the blockage structure in its initial state is shown in Fig. 1(a), where the internal block prevents the release groove from being exposed to the drug. The internal block is rigidly attached to both the nitinol wire and the fixed end of the catheter using solid filler. This ensures that the section between the release groove and the fixed end does not undergo axial torsional deformation under any conditions, while remaining mechanically decoupled from the catheter wall.
\par
To facilitate the switching of the MCR between the two deformation modes, a 6-DoF robotic arm is employed to manipulate the EPM and generate the required magnetic field as shown in Fig. 1(b). By reorienting the applied magnetic field, the system can selectively generate either a bending moment or a torsional torque. This capability is demonstrated in the surgical scenario shown in Fig. 1(c) and (d). An external propulsion mechanism, attached to the proximal end of the catheter, provides translation, while a controllable bending moment steers the distal tip towards the drug-delivery target (indicated by the yellow dot in Fig. 1(b)). Once at the target, zero bending moment but a torsional torque is applied to the tip, preparing the MCR for the drug release operation. In each mode, the total bending or torsional torque applied to the MCR is denoted as $\bm{M}$, which equals to the sum of torque exerted on each IPM, represented by $\bm{m_1}$,$\bm{m_2}$, and $\bm{m_3}$.

\par
In this study, we use a collaborative arm positions a 4 kg EPM to generate magnetic fields of controllable direction and strength around the catheter, given by:
\begin{equation}
\bm{B} = \frac{\mu_0 \boldsymbol{\mu_e} \left( 3 \hat{\bm{r}} \hat{\bm{r}}^T - \bm{I} \right)}{4\pi \|\bm{r}\|^3}
\end{equation}
where $\bm{r}$ denotes the position vector relative to the EPM and IPM, $\hat{\bm{r}}$ represents the unit vector of $\bm{r}$, $\bm{I}$ is the identity matrix, $\mu_0$ is the vacuum permeability, which is equal to $4\pi \times 10^{-7}~\mathrm{N \cdot A^{-2}}$ and $\boldsymbol{\mu_e}$ is the magnetic moment of the EPM. 

\begin{figure*}[!t]
    \centering
    \includegraphics[width=1.0\linewidth]{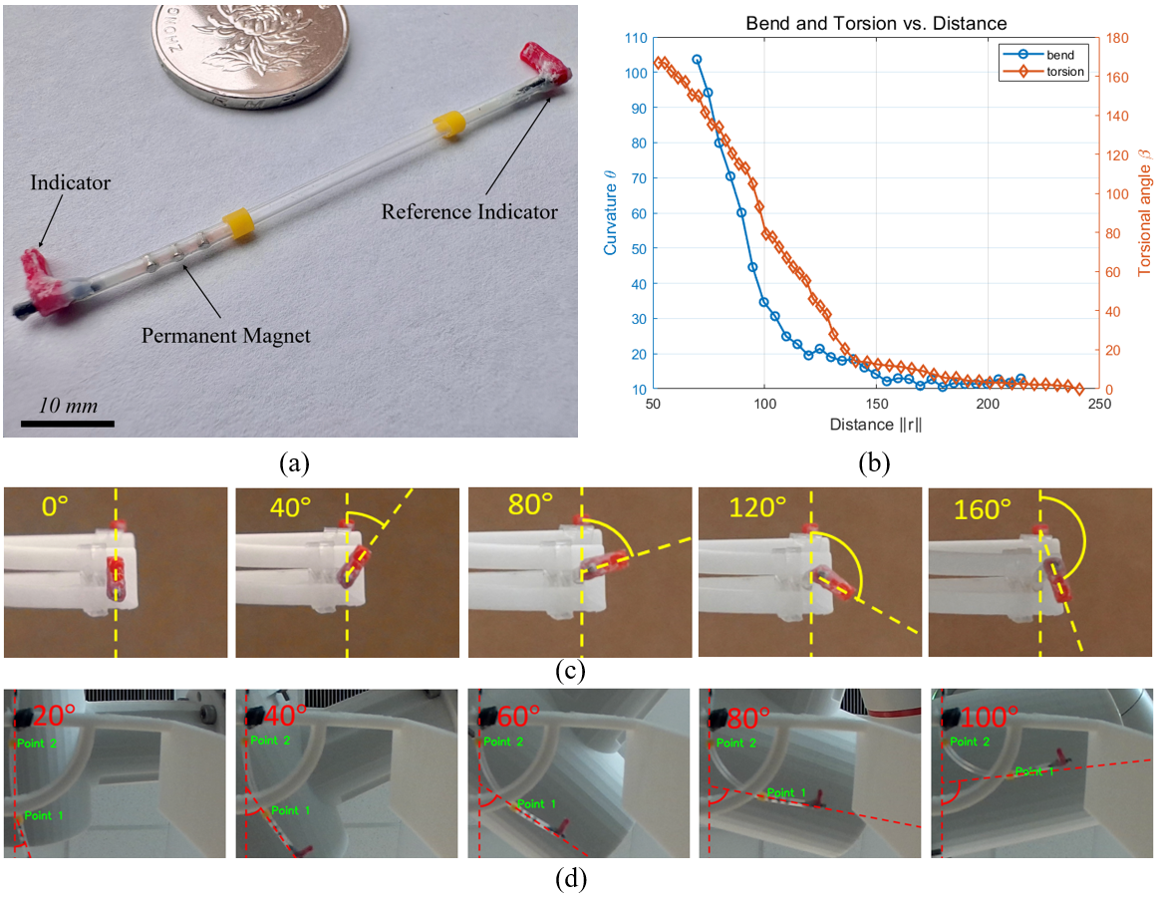}
    \caption{(a) Kinematic-prototype MCR (b) Record of curvature and torsional angle of the kinematic-prototype MCR (c) Capture of a series of recorded torsional angle (d) Capture of a series of recorded curvature.}
    \label{fig:2}
\end{figure*}
\par
The actuation principles for the bending and torsion modes are summarized in Fig. 1(c) and Fig. 1(d). In Fig. 1(c)(i), the applied magnetic field $\bm{B}$ is aligned with the distal axis of the catheter, so the shaft deforms within the bending plane. This alignment suppresses torsional torque and increases $\alpha$ to generate the required bending moment. Fig. 1(c)(ii) shows the view in the bending plane: $\mu_1$, $\mu_2$ and $\mu_3$ are the magnetic moments of the three IPMs, the yellow dot indicates their centroid, and $\theta$ is the distal bending angle. Let $\theta_i$ be the local curvature at each IPM location; the torque on the \textit{i-th} IPM is:

\begin{equation}
\bm{m_i} = \frac{\mu_0 \|\boldsymbol{\mu}\| \|\boldsymbol{\mu_e}\| sin\left( \alpha-\theta_i-\frac{\pi}{2} \right)}{2\pi \bm{r}^3}
\end{equation}
\par
According to [18], by denoting the direction of the \textit{c-th} IPM with $\theta_c$, the tip curvature can be described as:
\begin{equation}
\theta = \sum_{i=1}^{3}\sum_{c=i}^{3}\frac{\mu_0 \|\boldsymbol{\mu}\| \|\boldsymbol{\mu_e}\| sin\left( \alpha-\theta_c-\frac{\pi}{2} \right)L}{2\pi \bm{r}^3EI_{\mathrm{ner}}}
\end{equation}
where L is the bendable length of the catheter between each IPM, and $EI_\mathrm{ner}$ represents the flexural rigidity of the catheter. 
\par
For the torsional mode shown in Fig. 1(d), the EPM are oriented coplanar with the MCR's cross-section. According to (1), in this configuration, the magnetic torque applied to the IPMs will always align with the MCR's central axis. The external magnetic field will attempt to rotate the IPMs to align with it, thereby causing the MCR to undergo torsion.
\par
As shown in Fig. 1(d), there are four steps for using the MCR to deliver drugs via torsion. In Step 1, with the MCR in a torque-free state, the release groove aligns with the internal block. In Step 2, as the distanc $\lVert\bm{r}\rVert$ decreases, the MCR twists while each IPM’s magnetic moment $\boldsymbol{\mu}$ rotates in response to $\bm{B}$. The Release Groove then disengages from the Internal Block and the tip is subjected to a sustained torque $\bm{M}$. The drug does not exit because the air segment separates the drug from the groove. In Step 3, a torsional torque $\bm{T}$ is applied at the other end of the catheter opposite to $\bm{M}$. The catheter exhibits pronounced torsion and the Release Groove rotates substantially, but release still does not occur because the deformation alone does not create a sufficient internal volume change. In Step 4, as $\bm{T}$ is further increased, the catheter coils, markedly reducing its internal volume and expelling the drug. 
\par
While the MCR experiences both magnetic torque 
$M$ and twist torque 
$T$, its external tube will twist while the internal block rotates rigidly with the fixed end. As a result, this action causes the release groove to slide away from the block. To analyze the distance the groove slides, we first divide the external tube into two segments with lengths $d_1$ and $d_2$, where $d_1$ denotes the portion between the distal tip and the release groove, and $d_2$ represents the segment from the groove to the fixed end. Both segments of the external tube exhibit the same rigidity, but the portion from the groove to the fixed end will experience friction due to relative slip between the tube wall and the filler when torque is applied to the MCR. Therefore, the internal moment of the catheter can be expressed as:
\begin{equation}
\begin{split}
\tau(x)
&=T+M-\int_{0}^{x} q(s)\,ds\\
&=\begin{cases}
T+M-qd_{2}, & 0\le x < d_{1},\\[4pt]
T+M-q\bigl(d_{1}+d_{2}-x\bigr),     & d_{1}\le x \le d_{2} .
\end{cases}
\end{split}
\end{equation}
\par
Here, \textit{x} denotes the axial distance from the catheter tip to the evaluation point, where 
$x=0$ is the distal tip and $x=d_1 + d_2$ is the fixed end. \textit{s} is the arc-length integration variable, and \textit{q} is the distributed resisting torque due to friction between the filler and the catheter wall. Using the angle-of-twist relation, the torsional rotation at any position along the catheter is:
\begin{equation}
\varphi(x)=\frac{1}{GJ}\int_{0}^{x} T(s)\,ds
\end{equation}
where \textit{G} is the shear modulus and \textit{J} is the polar moment of inertia. Because each cross-section is annular, \textit{J} can be taken as constant. After specifying \textit{J}, evaluating \textit{$\varphi$} at the Release Groove and at the fixed end, and taking the difference, the relative twist between the Internal Block and the Release Groove follows as:
\begin{equation}
\varphi(d_2) - \varphi(d_1)=\frac{d_{2}}{GJ}\!\left[T+M-\frac{qd_{2}}{2}\right]
\end{equation}
\par
Thus, the sliding distance between the groove and the internal block is $d_2 \cdot tan(\varphi(d_2) - \varphi(d_1))$, Therefore, when fabricating the MCR, it is crucial to provide the internal block with a sufficiently long $d_2$, which ensures an adequate distance between the groove and the block under the applied twist torque.
\par
To sum up, this system supports two deformation modes with complementary roles in intervention. In the bending mode, an axial magnetic field steers the distal tip to follow tortuous anatomy, enabling lumen tracking, obstacle avoidance, and precise pointing to a target site. In the torsion mode, a coplanar field generates controlled twist that drives the dual-layer blocking mechanism; this twist selectively opens the outlet to realize on-demand, site-specific drug release while maintaining positional stability.

\section{Simulation and Experimental Validation}
According to the proposed MCR design, simulation and two experiments were implemented to test its performance and function.

\subsection{Deformation Ability Evaluation}
First, to verify that the two deformation modes are both achievable and decoupled under practical magnetic fields, finite-element analysis in ANSYS was performed, confirming that radial and axial magnetic torques produce the intended torsion and bending, respectively. Next, a catheter segment 80 mm in length was fabricated, with its proximal end rigidly clamped and three IPMs embedded radially at the distal tip. As shown in Fig. 2(a), a red indicator is placed at each end to mark the direction of the magnetic moments of the three IPMs. Under zero applied torque, the two indicators remain aligned. In addition, two yellow labels were affixed along the shaft to provide coordinate references of specific points on the MCR. These points can be extracted after HSV thresholding is applied to isolate the yellow regions in the captured images.
\par
For each actuation mode, after the EPM was positioned and the separation $\bm{r}$ was fine-tuned, the resulting MCR deformations were recorded. Representative torsional outcomes are shown in Fig. 2(c), where the camera view is aligned with the catheter’s central axis. The holding mechanism left space for distal bending by supporting the MCR near each Indicator in the vertical direction while rigidly clamping the Reference Indicator side. By manually annotating the direction of each Indicator, the angle between the annotated lines corresponds to the relative angle between the proximal and distal tips of the 80 mm sample. Fig. 2(c) confirms that fine-tuning the separation $\bm{r}$ enables precise torsional control.
\par
In the bending test, the EPM was positioned and oriented as illustrated in Fig. 1(c) using a 6-DoF robotic arm. The catheter curvature was calculated from the 3D coordinates of the two yellow labels, extracted by a stereo camera calibrated to the robotic-arm base frame. Representative bending results are provided in Fig. 2(d), demonstrating that curvature can also be accurately controlled by adjusting $\bm{r}$. The measured curvature $\theta$ and torsional angle $\beta$ as functions of $\|\bm{r}\|$ are summarized in Fig. 2(d), showing maximum bending and torsional angles of $105^\circ$ and $162^\circ$, respectively.

\begin{figure}[!t]
    \centering
    \includegraphics[width=1\linewidth]{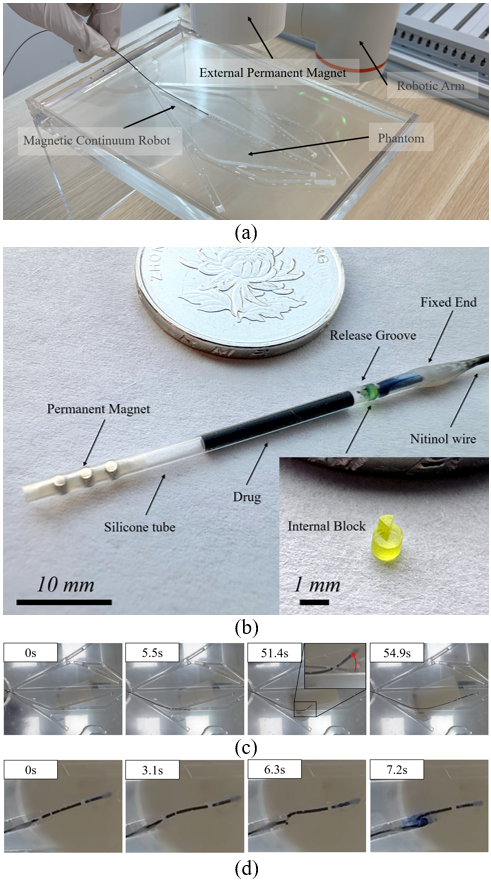}
    \caption{(a) Overview of the experiment platform (b) Fabricated prototype of MCR (c) Key frame of navigation by using bending mode (d) Key frame of drug delivery by using torsional mode.}
    \label{fig:3}
\end{figure}

\subsection{In-Phantom Intervention Experiment}
This section evaluates the MCR introduced in Fig. 1(a) through a benchtop experiment designed to emulate in-vivo conditions. The experimental platform, shown in Fig. 3(a), consists of a PDMS-based anatomical phantom mounted in an acrylic tank containing multiple vessel-like channels. Each channel has a square cross-section with a width of 5 mm. In total, seven distinct routes are included, with lengths ranging from 110 mm to 205 mm and intersection angles at junctions varying between $10^\circ$ and $50^\circ$. Both the tank and the channels are filled with water to approximate intraluminal conditions. The MCR is actuated by an EPM mounted on a robotic arm, while axial feed and proximal torsion are applied manually. The experimental task is to guide the MCR to the distal end of a chosen channel and release drug at the designated target. Following the design described earlier, the fabricated prototype is shown in Fig. 3(b). In this trial, ink is used as the drug substitute, and the Internal Block is fabricated with a micro-nano 3D printer. The block consists of a cylinder with a radius of 0.75 mm and height of 1 mm, topped by a cylindrical sector of the same radius, with a height of 0.5 mm and a central angle of $60^\circ$.

\par
Seven intervention–release trials were carried out, each targeting a different branch. In all cases, the MCR successfully reached the distal end of the selected branch and released the drug. Supplemental videos document the full procedures, while representative key frames are presented in Fig. 3(c) and Fig. 3(d). The navigation sequence in Fig. 3(c) illustrates the MCR entering the phantom, passing a junction, selecting the intended branch, and reaching the target. Time stamps highlight the intervals between successive frames. From entry into the phantom to arrival at the target, the procedure took 54.9 seconds in this example. Traversing between two intersections required about five seconds, but the majority of time, which was approximately 45 seconds, was spent positioning and orienting the EPM for branch selection. As shown in the third frame of Fig. 3(c), at 51.4 seconds the applied magnetic field was sufficiently strong to induce a pronounced bend of the MCR tip, enabling successful branch selection and target lumen access.
\par
The drug-release sequence is summarized in Fig. 3(d) with four close-up frames and their time stamps. In the first frame, the EPM is re-oriented manually correctly positioned close to the MCR so that the magnetic moments of the IPMs align upward. Next, by twisting the nitinol wire next to the fixed end manually, torque $\bm{T}$ is applied at the proximal end and gradually increased. At 3.1 s, slight deformation appears. At 6.3 s, the catheter exhibits pronounced helical deformation, yet no drug is released. At 7.2 s, the catheter coils tightly, expelling the internal air segment together with the drug. However, a single torsion actuation does not produce a large dose. If the wire side is loosen from hand at this point, $\bm{T}$ will be reduced to zero after the initial release. The catheter will elastically return toward its original shape and draw surrounding fluid through the Release Groove, which mixes with the residual drug. Then, applying torsion again releases a more diluted dose. Repeating this cycle progressively lowers the internal drug concentration until it approaches that of the surrounding fluid.
\section{Discussion}
Several experimental observations merit further discussion. 
\subsection{Deformation Result Reflection}
In Fig. 2(b), two curves exhibit noticeable jitter. For the blue curve, representing the bending angle $\theta$ of the MCR versus distance $\|\bm{r}\|$, fluctuations are most evident when $\|\bm{r}\|$ is between 100 mm and 150 mm. This is due to the curvature being calculated from two reference points after HSV-based image processing. As a result, visual noise is introduced, which affects the accuracy of the measurements.
\par
Additionally, both the bending and torsional curves deviate from linear trends with respect to distance. This is expected because, according to equation (2), the magnetic field $\bm{B}$ decreases with the cube of the distance $\|\bm{r}\|$, causing the applied torque to increase steeply as the EPM approaches the MCR. As a result, when $\|\bm{r}\|$ is greater than 150 mm, the bending curve shows minimal change. However, when $\|\bm{r}\|$ is less than 100 mm, the bending curve increases significantly, with the recorded maximum bending angle reaching $105^\circ$.
\par
For the torsional curve, the slope increases notably when $\|\bm{r}\|$ ranges from 80 mm to 130 mm, while other segments remain flatter. A clear bias appears around a torsional angle of $90^\circ$, caused by the catheter not remaining fully straight during twisting. Specifically, at about $90^\circ$, the distal tip retracted slightly and the middle segment lost tension, leading to a plateau until $\|\bm{r}\|$ decreased further and the torsional angle rose sharply. This effect is likely due to the EPM being placed too close to the IPMs, applying both torque and magnetic force. However, this artifact would not occur in practical use, as the distal tip of the MCR would be constrained by its surrounding environment.
\par
Overall, these findings confirm that the proposed MCR is capable of accurate and independent bending and torsional control through its two distinct actuation modes.
\subsection{Drug-Release Process Reflection}
In the in-phantom experiment, it was observed that twisting the MCR solely through magnetic torque to misalign the Release Groove and the Internal Block did not result in drug release. This occurred because the drug-carrying segment of the MCR experienced no significant change in the volume of its internal cavity. When additional manual torsion was applied at the proximal end, however, more pronounced deformation compressed the cavity, thereby squeezing out the drug. Conversely, if only proximal twisting is applied without an external magnetic field acting on the distal IPMs, the distal tip is not constrained; the entire catheter rotates together with the proximal side under torque. In this case, there is no relative displacement between the Internal Block and the Release Groove, even when the nitinol wire is continuously twisted. These observations indicate that the combined control strategy proposed in this study is both reasonable and effective.
\section{CONCLUSIONS}
This work presents a magnetic continuum robot capable of reliably switching between bending and torsion using a single external permanent magnet (EPM) and a simple radial magnet-embedding strategy. By reorienting the applied magnetic field, the system generates either a pure bending moment or a controlled twisting torque, while keeping all actuation hardware external to the anatomical region. A twist-triggered, dual-layer blockage mechanism further converts torsional shear into on-demand, site-specific drug release without the need for onboard valves or tethers.
\par
Modeling and analysis established the relationship between the applied field and resulting torque, as well as the tip response, including the relative twist and sliding distance between the internal block and the release groove. Finite element simulations confirmed that axial magnetic fields induce bending, while coplanar fields generate torsion, as intended. Benchtop characterization using a stereo camera quantified deformation maps of $\theta$ and $\beta$ versus $\|\bm{r}\|$, revealing maximum bending and torsional angles of $105^\circ$ and $162^\circ$, respectively. Experiments in a PDMS-based anatomical phantom validated the effectiveness of the proposed MCR for lumen following via bending and twist-activated drug release.
\par
Together, these results indicate that the proposed design offers a compact solution for achieving multifunctional deformation and targeted drug delivery. The dual-mode actuation enables long-range navigation and local task execution within tortuous pathways, positioning MCRs as promising candidates for minimally invasive, site-specific therapies.
\par
In future work, we plan to develop an automatic mode-switching algorithm to accelerate the reorientation phase of EPM manipulation, thereby reducing the operational time of the MCR. Additionally, we will explore alternative MCR designs that leverage the proposed dual deformation mode to improve release efficiency under torsional actuation (e.g., optimized groove/plate geometries, compliant reservoirs, and low-loss flow paths), with the goal of achieving higher delivery efficiency and making sure the released drug in target position stays at a high concentration even in a blood circulatory system.

\addtolength{\textheight}{-4cm}   




\end{document}